# ACSGRegNet: A Deep Learning-based Framework for Unsupervised Joint Affine and Diffeomorphic Registration of Lumbar Spine CT via Cross- and Self-Attention Fusion

XIAORU GAO

Institute of Medical Robotics, Shanghai Jiao Tong University

GUOYAN ZHENG*

Institute of Medical Robotics, Shanghai Jiao Tong University

Registration plays an important role in medical image analysis. Deep learning-based methods have been studied for medical image registration, which leverage convolutional neural networks (CNNs) for efficiently regressing a dense deformation field from a pair of images. However, CNNs are limited in its ability to extract semantically meaningful intra- and inter-image spatial correspondences, which are of importance for accurate image registration. This study proposes a novel end-to-end deep learning-based framework for unsupervised affine and diffeomorphic deformable registration, referred as ACSGRegNet, which integrates a cross-attention module for establishing inter-image feature correspondences and a self-attention module for intra-image anatomical structures aware. Both attention modules are built on transformer encoders. The output from each attention module is respectively fed to a decoder to generate a velocity field. We further introduce a gated fusion module to fuse both velocity fields. The fused velocity field is then integrated to a dense deformation field. Extensive experiments are conducted on lumbar spine CT images. Once the model is trained, pairs of unseen lumbar vertebrae can be registered in one shot. Evaluated on 450 pairs of vertebral CT data, our method achieved an average Dice of 0.963 and an average distance error of 0.321mm, which are better than the state-of-the-art (SOTA).

## 1 INTRODUCTION

Nowadays, spine surgeries, such as surgical resection of spinal column tumors and disc replacement surgery are increasingly dependent on image guidance [1]. However, it is still challenging to integrate this technique into surgery effectively because the interpretation of intraoperative images is time-consuming. Therefore, pre-operative images like computed tomography (CT) are needed to assist surgery procedures. In order to detect quantitatively changes between images at different time or of different subjects, it is essential to establish a semantically meaningful mapping between them by image registration.

___________________________________

*Corresponding author: Prof. Dr. Guoyan Zheng (Email: guoyan.zheng@sjtu.edu.cn).

Image registration is the determination of geometrical transformation that aligns objects in both given images. It benefits clinical diagnosis, treatment planning, and computer assisted inventions of various diseases. In this study, we are aiming to address the image registration problem for the spine. Unlike other bony structures, the spine is a complex structure and would vary greatly between patients [2]. A lot of work has been done aiming to address spine registration problems. Gill et al. [3] took each lumbar vertebra in CT as a sub-volume and transformed them individually. Hille et al. [4] split a global non-rigid registration task into multiple local rigid registration. Glocker et al. [5] used a classifier to estimate the locations of vertebrae and integrated these location priors into registration method. However, these conventional registration methods are time-consuming, which is difficult to be used for intra-operative interventions.

Recently, with the development of deep learning (DL) techniques, many researchers have turned to DL-based methods to achieve fast registrations [6, 7, 8, 9]. Although supervised learning methods can get more accurate registration results, the labeled data are not always trivial to obtain. Unsupervised methods such as VoxelMorph [9] and its variants [7, 8] need pre-aligned data to predict the deformation field, which is usually obtained via an affine registration. One of the common limitations, however, lies in the fact that these methods ignore the intra- and inter-image spatial feature relevance by simply concatenating the image pair as the input to a CNN network, leading to failure in finding semantically meaningful correspondences. In this paper, we propose an unsupervised end-to-end DL-based network, referred as ACSGRegNet, which integrating a cross-attention module for establishing inter-image feature correspondences and a self-attention module for intra-image anatomical structures aware. Both attention modules are built on transformer encoders. The output from each attention module is respectively fed to a decoder to generate a velocity field. We further introduce a gated fusion module to fuse both velocity fields. The fused velocity field is then integrated to a dense deformation field. We hypothesize that by converting image features to a spatial relationship, we can get better results on registration of lumbar spine CT images than the state-of-the-art (SOTA).

## 2 METHOD

Our method is divided into two main parts, with an optional auxiliary information part. The complete framework can be trained end-to-end. The first part is an affine registration network, which estimates an affine transformation. The second part is a diffeomorphic deformable registration network which accounts for local deformations. The overall framework is illustrated in Figure 1. Let $f$ and $m$ denote the given pair of fixed and moving images, respectively. The input to the framework is the concatenation of $f$ and $m$, and the output is the warped moving images after affine and deformable registration. The segmentation masks of the vertebrae can be leveraged as auxiliary information by propagating the moving image segmentation masks during training. We will elaborate on different parts as follows.

### 2.1 Network Architecture

The affine registration network architecture is depicted in Figure 2. It takes the concatenated $f$ and $m$ as the input and outputs 12 affine transformation parameters. Each ConvBlock is defined as the combination of a 3D convolutional layer (Conv) with a kernel size of $3\times3\times3$, a Batch Normalization layer (BN) and a LeakyReLU activation function. A total of five downsampling ConvBlocks with a stride of 2 are used to broaden receptive fields. Each of them is followed by two ConvBlocks with a stride of 1. The affine parameters are then computed by a fully connected layer.

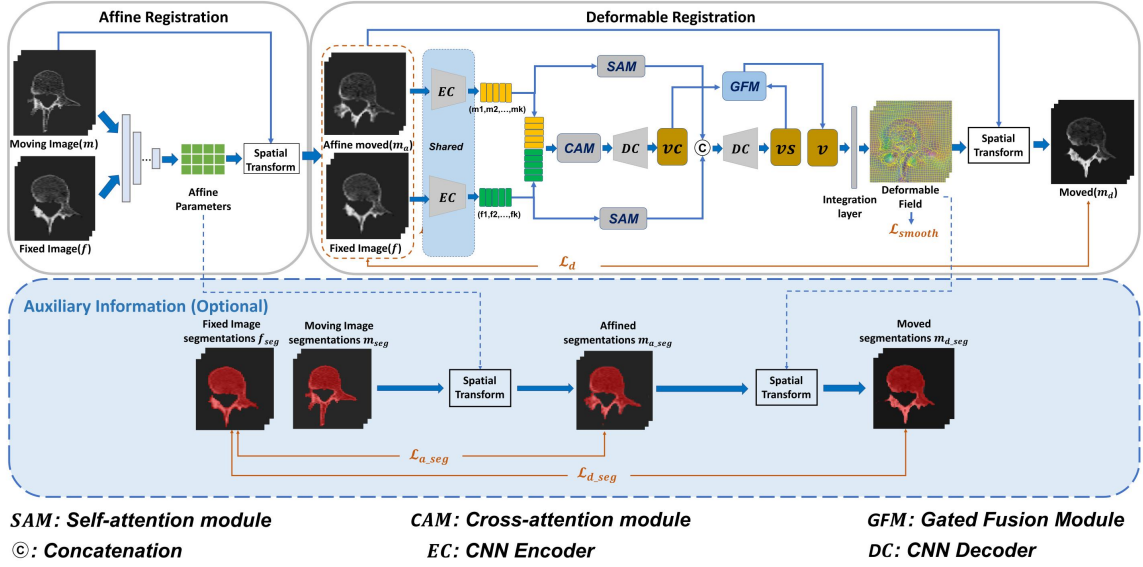

Figure 1: A schematic illustration of the network architecture of the proposed method. It consists of an affine (left) and a diffeomorphic deformable (right) registration component. The affine part learns 12 affine parameters and outputs the affinely warped moving image $m_a$. Then the fixed image $f$ and the affinely registered image $m_a$ are subsequently inputted into the self- and cross-attention modules and CNN decoders to estimate two velocity fields $vs$ and $vc$. A gated fusion model (GFM) is proposed to fuse the two velocity fields to get the final velocity field $v$ which can then be used to generate the final deformation field via an integration layer. The finally warped moving image $m_d$ is obtained by warping $m_a$ with the spatial transform using the integrated deformation field. Auxiliary information such as vertebral segmentations $f_{seg}$, $m_{seg}$ can be leveraged during training (optional blue box). Losses are marked in orange.

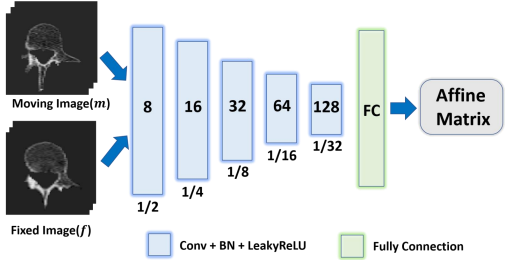

Figure 2: Affine registration network architecture.

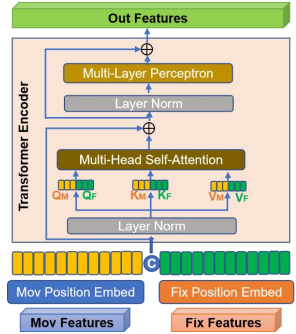

Figure 3: Architecture of the transformer encoder module (**TEM**).

Next, given an image pair of affinely registered image $m_a$ and $f$, we first use a CNN-based encoder (EC) to extract high-level embeddings of both images, which are then collapsed into k-dimensional (we empirically chose k=252) vector sequences $\widetilde{E}_{m_a} = \{m_1; m_2; ...; m_k\}$ and $\widetilde{E}_f = \{f_1; f_2; ...; f_k\}$, respectively with a trainable linear projection. Then, we utilize a self-attention module to exploit the intra-image feature enhancement and a cross-attention module to model both the inter- and intra-image correspondences. Both of them are built on a transformer encoder module (TEM) as shown in Figure 3. Below we present details of each module.

### 2.1.1 Self-attention module (SAM)

The proposed SAM consists of 12 layers of vanilla transformer encoder [10], as shown in Figure 3. The sequence embeddings $\widetilde{E}_{m_a}$ and $\widetilde{E}_f$ are respectively added with the learnable position embeddings to get $E_{m_a}$ and $E_f$, which are used as the inputs to the SAM to extract intra-image position and context dependent features. In total there are two SAMs, one for the fixed image and the other for the affinely registered moving image. The query, key and value are formed with following equations:

$$K_i = E_i W_i^K; \quad Q_i = E_i W_i^Q; \quad V_i = E_i W_i^V; \text{ where } i \in \{m_a, f\} \tag{1}$$

Where $W_i^K$, $W_i^Q$ and $W_i^V$ are learnable weight matrices.

The self-attention is calculated as:

$$Attention\ (Q_i, K_i, V_i) = \text{softmax}\left(\frac{Q_i K_i^T}{\sqrt{k}}\right) V_i \tag{2}$$

Thereafter, we concatenate the output features from the two SAMs as the input to a CNN decoder (DC) to regress a velocity field $vs$.

### 2.1.2 Cross-attention module (CAM)

The proposed CAM also consists of 12 layers of vanilla transformer encoder [10], as shown in Figure 3. After adding position embeddings, we stack the result $E = \begin{pmatrix} E_f \\ E_{m_a} \end{pmatrix}$ as the input to the CAM to extract both intra- and inter-image position and context dependent features. Here, the query, key and value are formed with following equations:

$$K_E = E W^K; \quad Q_E = E W^Q; \quad V_E = E W^V; \tag{3}$$

Where $W^K$, $W^Q$ and $W^V$ are learnable weight matrices.

The cross-attention is calculated as:

$$Attention\ (Q_E, K_E, V_E) = \text{softmax}\left(\frac{Q_E K_E^T}{\sqrt{k}}\right) V_E \tag{4}$$

The output features from the CAM are taken as the input to a CNN decoder to get another velocity field $vc$.

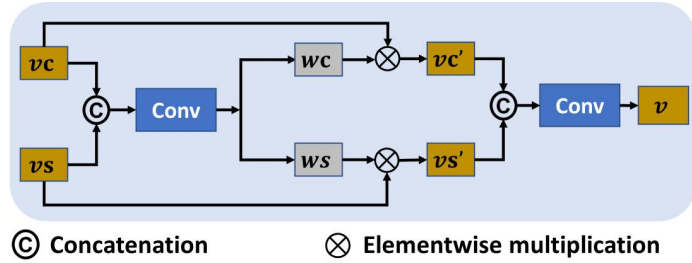

Figure 4: The proposed gated fusion attention module (*GFM*) for fusing two velocity fields. $vc$ and $vs$ represents the two velocity fields from two different decoders, respectively. $wc$ and $ws$ are the gated attention weights. $vc'$ and $vs'$ are the weighted results, respectively. $v$ is the final velocity field that is used to generate the final deformation field via an integration layer.

### 2.1.3 Gated fusion of velocity fields

We further propose a Gated Fusion Module (GFM) to fuse the two velocity fields $vs$ and $vc$. Figure 4 shows the architecture of the proposed GFM. The two velocity fields are concatenated before passing to a convolution layer. The two weight matrices are then used to multiply with the original velocity fields. The weighted velocity fields are concatenated again and pass into another convolution layer to obtain the final velocity field, which is used to integrated for the deformation field.

### 2.1.4 Implementation details

Our CNN encoder (EC) and decoder (DC) are part of a 3D U-Net stacked with the proposed self- and cross-attention modules. Shared encoder path is used for obtaining volumetric image embeddings while different decoder paths are used to infer the velocity fields. Skip connections are added between the corresponding layers of the encoder and decoder paths.

## 2.2 Loss Function

The loss function consists of three terms. First is a similarity loss defined by local normalized cross-correlation ($LNCC$) between the warped moving images and the fixed images in both affine and deformable registration ($\mathcal{L}_a$ and $\mathcal{L}_d$), respectively. $LNCC$ is defined by:

$$LNCC(f, w) = \sum_{\mathbf{p} \in \Omega} \frac{\left(\sum_{\mathbf{p}_i}(f(\mathbf{p}_i) - \hat{f}(\mathbf{p}))(w(\mathbf{p}_i) - \hat{w}(\mathbf{p}))\right)^2}{\left(\sum_{\mathbf{p}_i}(f(\mathbf{p}_i) - \hat{f}(\mathbf{p}))^2\right)\left(\sum_{\mathbf{p}_i}(w(\mathbf{p}_i) - \hat{w}(\mathbf{p}))^2\right)} \quad (5)$$

where $f$ and $w$ indicate the fixed and warped images, respectively, with $\hat{f}(\mathbf{p})$ and $\hat{w}(\mathbf{p})$ represent the mean values computed from a sub-volume of size $n^3$ around voxel $\mathbf{p}$. Therefore we have:

$$\mathcal{L}_a = -LNCC(f, m_a) \text{ and } \mathcal{L}_d = -LNCC(f, m_d) \quad (6)$$

Second is a diffusion regularizer $\mathcal{L}_{smooth}$ on the spatial gradients of the estimated deformation field $\phi$ along each axis:

$$\mathcal{L}_{smooth} = \sum_{p \in \Omega} \|\partial \phi_x(p)\|^2 + \|\partial \phi_y(p)\|^2 + \|\partial \phi_z(p)\|^2 \quad (7)$$

Third is an auxiliary Dice loss when segmentation masks of both fixed and moving images are available:

$$\text{Dice} = \frac{2|f_{seg} \cap w_{seg}|}{|f_{seg}| + |w_{seg}|} \quad (8)$$

where $f_{seg}$ is the fixed image segmentation mask and $w_{seg}$ is the warped image segmentation mask. It is evaluated in affine and deformable registration ($\mathcal{L}_{a\_seg}$ and $\mathcal{L}_{d\_seg}$), respectively. Finally, the total loss function is defined as the weighted sum of the terms defined above:

$$\mathcal{L} = \lambda_1 \mathcal{L}_a + \lambda_2 \mathcal{L}_d + \lambda_3 \mathcal{L}_{smooth} + \lambda_4 \mathcal{L}_{a\_seg} + \lambda_5 \mathcal{L}_{d\_seg} \quad (9)$$

where $\lambda_1$, $\lambda_2$, $\lambda_3$, $\lambda_4$ and $\lambda_5$ are weighting coefficients controlling the relative weights of different terms. Empirically, we chose $\lambda_1$ = 0.3, $\lambda_2$ = 0.7, $\lambda_3$ = 0.001, $\lambda_4$ = 0.01 and $\lambda_5$ = 0.1.

## 3 EXPERIMENT

### 3.1 Dataset and Preprocessing

The dataset consists of 61 CT scans of lumbar vertebrae with their corresponding segmentation masks. All scans are different in orientation, size and spacing. We first reoriented all scans and segmentation masks before resampling them to a unified voxel size of $1 \times 1 \times 1$ mm³. Next we multiplied the image data with their segmentation masks to obtain individual lumbar vertebra before cropping them to a size of $128 \times 128 \times 64$. Finally, for the five lumbar vertebrae L1-L5, we obtained 58 L1, 60 L2, 61 L3, 61 L4, and 60 L5, respectively.

In this paper, we train the model for inter-subject registration. For each level of lumbar vertebra, we randomly choose 10 samples for testing and the rest samples for training. As each sample can be used as fixed image with respect to others, we obtained 2256 pairs of L1, 2450 pairs of L2, 2550 pairs of L3, 2550 pairs of L4 and 2450 pairs of L5 as the training dataset. The testing dataset size is 90 pairs for each level of lumbar vertebra. As we have the segmentation masks for all the dataset, we evaluate the performance of each trained model with four commonly used metrics, i.e., Dice coefficient, Precision (PREC), Recall (REC), average symmetric surface distance (ASSD) between the fixed image segmentation mask and the warped moving image segmentation mask. Additionally, we compute the number and the percentage of voxels with non-positive Jacobian determinant (i.e., $|J_\phi| \leq 0$ and % mean of $|J_\phi| \leq 0$) to quantify the deformation regularity.

### 3.2 Comparison with the state-of-the-art (SOTA)

We compared the proposed method with two SOTA deep learning-based registration methods, i.e., VoxelMorph [9] and ViT-V-Net [11]. As there is no affine registration part for both previous works, we added affine registration part to both networks and referred the networks with affine registration as AVoxelMorph and AViT-V-Net, respectively. We trained all three networks on the same dataset with mixed levels of lumbar vertebrae. Specifically, we sampled 10 pairs for each sample of each level vertebra from the training dataset and obtain a total number of 2500 pairs as the training dataset. We additionally sample 1 pair for each sample of each level vertebra from the training dataset and obtain a total number of 250 pairs as the validation dataset. Once the models were trained, any pair of unseen lumbar vertebrae could be registered. Performance of the three trained models were evaluated on the same testing dataset. All three models were trained end-to-end on a NAVIDIA TITAN RTX GPU, using an Adam optimizer starting at a learning rate of $10^{-5}$ with a minibatch size of 8.

### 3.3 Ablation study

We designed and conducted three ablation studies to investigate the effectiveness of the proposed cross- and self-attention modules (CAM and SAM), as well as the gated fusion module (GFM), on the overall performance of our proposed registration method. Our baseline network is the proposed ACSGRegNet after removing all three modules, which is referred as BaseModel. The second model is obtained by adding SAM to the BaseModel while the third model is derived by adding CAM to the BaseModel. Finally, we obtain the proposed ACSGRegNet by adding all three modules to the BaseModel. All four models are then trained on the same training dataset, and evaluated on the same testing dataset.

Table 1. Performance of AVoxelMorph when evaluated on the testing dataset.

| Vertebra | Dice | | | PREC | | | REC | | | ASSD(mm) | | |
|---|---|---|---|---|---|---|---|---|---|---|---|---|
| | Initial | Affine | Final | Initial | Affine | Final | Initial | Affine | Final | Initial | Affine | Final |
| L1 | 0.610 | 0.669 | 0.938 | 0.627 | 0.597 | 0.931 | 0.627 | 0.788 | 0.945 | 2.879 | 2.453 | 0.480 |
| L2 | 0.610 | 0.731 | 0.954 | 0.617 | 0.656 | 0.946 | 0.617 | 0.833 | 0.961 | 2.915 | 2.064 | 0.381 |
| L3 | 0.642 | 0.744 | 0.959 | 0.650 | 0.676 | 0.951 | 0.650 | 0.837 | 0.967 | 2.986 | 2.053 | 0.365 |
| L4 | 0.623 | 0.744 | 0.957 | 0.634 | 0.661 | 0.954 | 0.634 | 0.852 | 0.960 | 3.196 | 2.142 | 0.380 |
| L5 | 0.629 | 0.691 | 0.930 | 0.637 | 0.612 | 0.925 | 0.637 | 0.803 | 0.936 | 3.061 | 2.615 | 0.635 |
| Average | 0.623 | 0.716 | 0.947 | 0.633 | 0.640 | 0.941 | 0.633 | 0.823 | 0.954 | 3.007 | 2.265 | 0.448 |

Table 2. Performance of AViT-V-Net when evaluated on the testing dataset

| Vertebra | Dice | | | PREC | | | REC | | | ASSD(mm) | | |
|---|---|---|---|---|---|---|---|---|---|---|---|---|
| | Initial | Affine | Final | Initial | Affine | Final | Initial | Affine | Final | Initial | Affine | Final |
| L1 | 0.610 | 0.699 | 0.906 | 0.627 | 0.645 | 0.870 | 0.627 | 0.782 | 0.949 | 2.879 | 2.202 | 0.749 |
| L2 | 0.610 | 0.743 | 0.930 | 0.617 | 0.690 | 0.904 | 0.617 | 0.811 | 0.960 | 2.915 | 1.950 | 0.595 |
| L3 | 0.642 | 0.745 | 0.934 | 0.650 | 0.702 | 0.909 | 0.650 | 0.801 | 0.961 | 2.986 | 2.033 | 0.604 |
| L4 | 0.623 | 0.763 | 0.938 | 0.634 | 0.699 | 0.915 | 0.634 | 0.845 | 0.963 | 3.196 | 1.979 | 0.590 |
| L5 | 0.629 | 0.690 | 0.893 | 0.637 | 0.645 | 0.864 | 0.637 | 0.752 | 0.927 | 3.061 | 2.557 | 0.969 |
| Average | 0.623 | 0.728 | 0.920 | 0.633 | 0.676 | 0.892 | 0.633 | 0.798 | 0.952 | 3.007 | 2.144 | 0.701 |

Table 3: Performance of the proposed ACSGRegNet when evaluated on the testing dataset

| Vertebra | Dice | | | PREC | | | REC | | | ASSD(mm) | | |
|---|---|---|---|---|---|---|---|---|---|---|---|---|
| | Initial | Affine | Final | Initial | Affine | Final | Initial | Affine | Final | Initial | Affine | Final |
| L1 | 0.610 | 0.691 | 0.958 | 0.627 | 0.632 | 0.950 | 0.627 | 0.776 | 0.965 | 2.879 | 2.242 | 0.329 |
| L2 | 0.610 | 0.738 | 0.966 | 0.617 | 0.672 | 0.960 | 0.617 | 0.825 | 0.972 | 2.915 | 1.995 | 0.280 |
| L3 | 0.642 | 0.762 | 0.969 | 0.650 | 0.699 | 0.962 | 0.650 | 0.843 | 0.976 | 2.986 | 1.912 | 0.278 |
| L4 | 0.623 | 0.760 | 0.967 | 0.634 | 0.701 | 0.965 | 0.634 | 0.835 | 0.969 | 3.196 | 1.976 | 0.291 |
| L5 | 0.629 | 0.703 | 0.954 | 0.637 | 0.638 | 0.947 | 0.637 | 0.793 | 0.962 | 3.061 | 2.457 | 0.425 |
| Average | 0.623 | 0.731 | 0.963 | 0.633 | 0.668 | 0.957 | 0.633 | 0.814 | 0.969 | 3.007 | 2.116 | 0.321 |

Table 4: The comparison on the number and percentage of voxels with a non-positive Jacobian determinant

| Vertebra | $|J_\phi| \leq 0$ | | | % mean of $|J_\phi| \leq 0$ | | |
|---|---|---|---|---|---|---|
| | AVoxelMorph | AViT-V-Net | ACSGRegNet | AVoxelMorph | AViT-V-Net | ACSGRegNet |
| | Mean | Mean | Mean | Mean | Mean | Mean |
| L1 | 27427 | 11931 | 10209 | 2.7 | 1.2 | 1.0 |
| L2 | 26938 | 13013 | 10532 | 2.7 | 1.3 | 1.0 |
| L3 | 29344 | 14729 | 11569 | 2.9 | 1.4 | 1.1 |
| L4 | 30017 | 14869 | 11181 | 3.0 | 1.5 | 1.1 |
| L5 | 37252 | 18524 | 13065 | 3.7 | 1.8 | 1.3 |
| Average | 30195 | 14613 | 11311 | 3.0 | 1.4 | 1.1 |

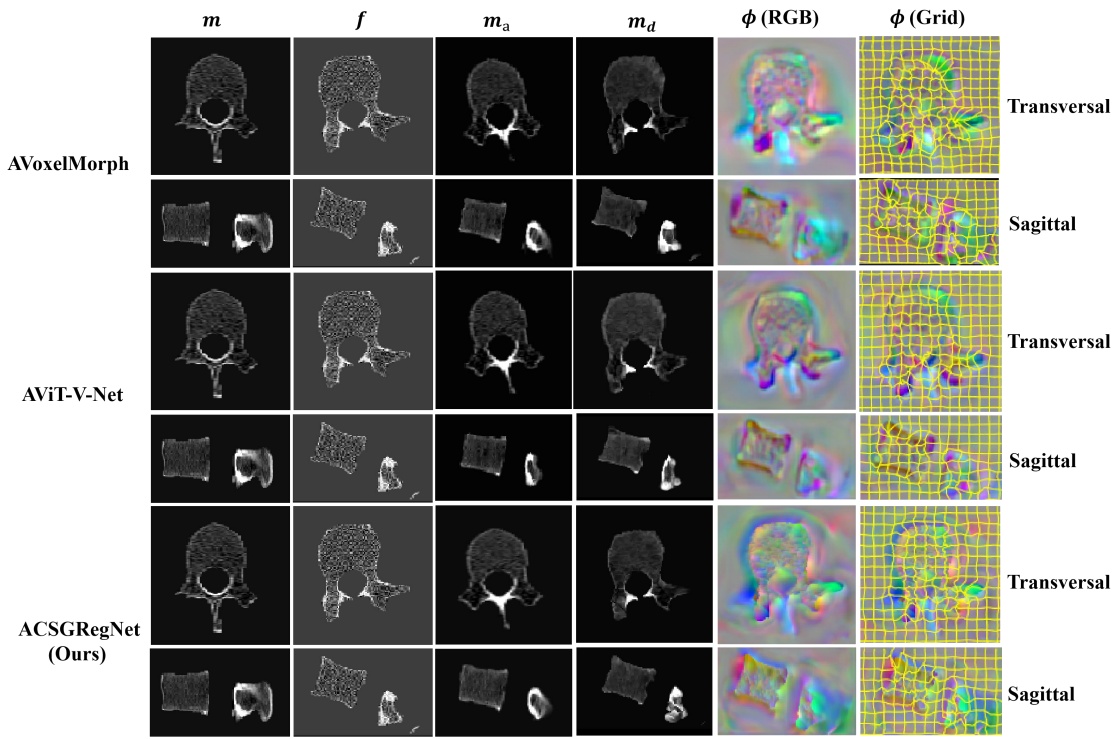

Figure 5: Qualitative comparison of all three methods. Columns 1-6 show the moving image $m$, the fixed image $f$, the affinely warped image $m_a$, the finally warped image $m_d$, the deformation fields displayed with color-coded visualization ($\phi$ (**RGB**)) and the deformation fields displayed with deformed grid ($\phi$(**Grid**)), respectively. Results are illustrated along middle transversal and sagittal planes. Top two rows: results obtained by the AVoxelMorph model; Middle two rows: results obtained by the AVit-V-Net model; Bottom two rows: results obtained by the proposed ACSGRegNet.

Table 5: Results of the ablation study

|  | Dice | | | PREC | | | REC | | | ASSD(mm) | | |
|---|---|---|---|---|---|---|---|---|---|---|---|---|
|  | Initial | Affine | Final | Initial | Affine | Final | Initial | Affine | Final | Initial | Affine | Final |
| BaseModel | 0.623 | 0.720 | 0.944 | 0.633 | 0.665 | 0.929 | 0.633 | 0.796 | 0.961 | 3.007 | 2.211 | 0.475 |
| BaseModel + SAM | 0.623 | 0.715 | 0.953 | 0.633 | 0.631 | 0.940 | 0.633 | 0.836 | 0.966 | 3.007 | 2.277 | 0.404 |
| BaseModel + CAM | 0.623 | 0.722 | 0.956 | 0.633 | 0.642 | 0.946 | 0.633 | 0.833 | 0.966 | 3.007 | 2.218 | 0.373 |
| The proposed method | 0.623 | 0.731 | 0.963 | 0.633 | 0.668 | 0.957 | 0.633 | 0.814 | 0.969 | 3.007 | 2.116 | 0.321 |

## 4 RESULTS

### 4.1 Results of the comparison study

Results of three models are shown in Table 1, Table 2, and Table 3, respectively. Table 1 shows the results of the AVoxelMorph model when evaluated on the testing dataset. For each metric, we show the results on three different stages, i.e., before registration (Initial), after affine registration (Affine) and finally after deformable registration (Final). As one can see from Table 1, AVoxelMorph achieved an average Dice coefficient of 0.947 and an average ASSD of 0.448mm. Table 2 presents the results of the AViT-V-Net. An average Dice coefficient of 0.920 and an average ASSD of 0.701mm are observed, which are worse than the AVoxelMorph model. Table 3 shows the results of our proposed ACSGRegNet. Evaluated on the same testing dataset, ACSGRegNet achieved an average Dice coefficient of 0.963 and an average ASSD of 0.321mm, which were much better than other two SOTA models. Table 4 further presents the number and the percentage of voxels with non-positive Jacobian determinant. As one can see from this table, the average number and percentage of voxels with non-positive Jacobian determinant obtained from the proposed ACSGRegNet are smaller than other two methods, demonstrating a better deformation regularity. Figure 5 shows qualitative comparison of three models.

## 4.2 Results of the ablation study

Results of the ablation study are presented in Table 5. As one can see from this table, the BaseModel achieved the worst results, with an average Dice coefficient of 0.944 and an average ASSD of 0.475mm. Adding SAM alone to the BaseModel, the average Dice coefficient is improved to 0.953 and the average ASSD is improved to 0.404mm. When only CAM is added to the BaseModel, we obtain an average Dice coefficient of 0.956 and an average ASSD of 0.373mm, demonstrating that both attentions are helpful to improve the registration performance and cross-attention module contributes more than the self-attention module. With the proposed ACSGRegNet, we obtain the best results, with an average Dice coefficient of 0.963 and an average ASSD of 0.321mm.

## 5 DISCUSSION

Inter-subject registration of lumbar spine CT images is challenging due to the large shape variations. This paper proposed a novel ACSGRegNet for unsupervised affine and diffeomorphic deformable registration, which integrated a cross-attention module for establishing inter-image feature correspondences and a self-attention module for intra-image anatomical structures aware. Both attention modules were built on transformer encoders. The output from each attention module was respectively fed to a decoder to generate a velocity field. We further proposed a gated fusion module to fuse both velocity fields. The fused velocity field is then integrated to a dense deformation field. We conducted comprehensive experiments to evaluate the efficacy of the proposed ACSGRegNet. Compared with the SOTA registration methods, the proposed ACSGRegNet demonstrated superior results, both quantitatively and qualitatively. We further designed and conducted ablation study to evaluate the effectiveness of each individual components of the proposed ACSGRegNet, demonstrating that all three modules are helpful to improve registration performance.

**ACKNOWLEDGMENTS**

The work was partially supported by the Natural Science Foundation of China via project U20A20199 and by Shanghai Municipal S&T Commission via Project 20511105205.